\documentclass[conference]{IEEEtran}
\IEEEoverridecommandlockouts

\usepackage[utf8]{inputenc} 
\usepackage[T1]{fontenc}    
\usepackage{hyperref}       
\usepackage{url}            
\usepackage{booktabs}       
\usepackage{amsfonts}       
\usepackage{nicefrac}       
\usepackage{microtype}      
\usepackage{algorithm,algpseudocode}
\usepackage{bbm}
\usepackage{hyperref}

\usepackage[table]{xcolor}

\usepackage{cite}
\usepackage{amsmath,amssymb}
\usepackage{dsfont}
\usepackage{graphicx}
\usepackage{textcomp}
\def\BibTeX{{\rm B\kern-.05em{\sc i\kern-.025em b}\kern-.08em
    T\kern-.1667em\lower.7ex\hbox{E}\kern-.125emX}}

\usepackage{amssymb}

\usepackage{amsmath}

\usepackage{mathtools}

\usepackage{lscape}
\usepackage{rotating}
\usepackage{amsmath}
\usepackage{amssymb}
\usepackage{mathtools}
\usepackage{amsthm}
\usepackage{multirow}
\usepackage{breqn}
\usepackage{caption}
\usepackage{subcaption}
\usepackage{tikz}
\usetikzlibrary{arrows, arrows.meta}
\usepackage{bbm}
\usepackage{adjustbox}
\usepackage{amssymb}
\usepackage{pifont}
\newcommand{\cmark}{\ding{51}}%
\newcommand{\xmark}{\ding{55}}%
\usepackage[capitalize,noabbrev]{cleveref}
\theoremstyle{plain}

\theoremstyle{definition}
\newtheorem{definition}{Definition}

\theoremstyle{remark}
\newtheorem{remark}{Remark}

\usepackage[textsize=tiny]{todonotes}

\begin{document}
\title{Quantifying Correlations of Machine Learning Models}

\author{
\IEEEauthorblockN{Yuanyuan Li}
\IEEEauthorblockA{\textit{Munich RE} \\
California, US  \\
yli@munichre.com}
\and
\IEEEauthorblockN{Neeraj Sarna}
\IEEEauthorblockA{\textit{Munich RE} \\
Munich, Germany \\
nsarna@munichre.com}
\and
\IEEEauthorblockN{Yang Lin}
\IEEEauthorblockA{\textit{Hartford Steam Boiler} \\
Connecticut, US \\
yang_lin@hsb.com}
}

\maketitle
\begin{abstract}
Machine Learning models are being extensively used in safety critical applications where errors from these models could cause harm to the user. Such risks are amplified when multiple machine learning models, which are deployed concurrently, interact and make errors simultaneously. This paper explores three scenarios where error correlations between multiple models arise, resulting in such aggregated risks. Using real-world data, we simulate these scenarios and quantify the correlations in errors of different models. Our findings indicate that aggregated risks are substantial, particularly when models share similar algorithms, training datasets, or foundational models. Overall, we observe that correlations across models are pervasive and likely to intensify with increased reliance on foundational models and widely used public datasets, highlighting the need for effective mitigation strategies to address these challenges.
\end{abstract}


\section{Introduction}
Machine Learning (ML) has experienced widespread adoption across a multitude of industries due to its potential to improve operational efficiency, enhance decision-making processes, and accelerate scientific discovery. From healthcare and finance to manufacturing and transportation, ML has driven significant technological advancements, automating tasks and unlocking new insights. However, as more organizations adopt ML models, it’s becoming clear that proper governance and risk management are essential \cite{AIGovernance}. With the growing complexity and number of models being deployed, understanding how these models interact—particularly how their errors may be linked—is becoming more important.

While the performance of individual models is often analyzed in isolation, the interplay between models and their simultaneous errors is an area that has received less attention. For instance, do errors in one model tend to coincide with errors in another model, especially when both models are applied to the same underlying problem or when they share common data sources or structures?

Although such questions haven't received much attention in the literature, study of error correlation between models is paramount in safety critical applications. Consider, for instance, medicine dosage prediction using AI \cite{radiationML}. Multiple hospital might develop dosage prediction models using similar public datasets. The underlying similarity of datasets would result in the errors of these models being correlated. Consequently, several of the hospital could simultaneously prescribe incorrect dosage to multitudes of patients, resulting in grave consequences. The domain of connected self-driving cars is similar where different technology providers develop their own models based upon similar underlying datasets \cite{MLAutonomousAdnan,MLAutonomousGourav,MLAutonomousQing}. 

The advent of fine-tune foundation models could further exacerbate the error correlations between models. Once trained, these foundation models can be fine-tuned for specific tasks by further training them on task-specific data, a process known as fine-tuning. A crucial question to then consider is: whether fine-tuned downstream models exhibit correlations and if their performance can be collectively influenced by the foundational model’s capabilities. This phenomenon, referred to as "homogenization" \cite{bommasani2021opportunities, Creel_Hellman_2022}, highlights that downstream models may converge toward a more uniform set of judgments rooted in the underlying foundation model. Consequently, any defects or biases present in the foundation model are likely inherited by all the fine-tuned models derived from it, amplifying systemic risks in AI applications. To the best of our knowledge, there has been no prior work that quantifies the extent of this effect. Our study addresses this gap by systematically analyzing and measuring the correlations among downstream models fine-tuned from the same foundation model.

The above discussion indicates that, especially for safety critical applications, error correlations can result in larger-scale disruptions that are difficult to predict and mitigate. Therefore, understanding and quantifying these error correlations is a critical step in developing robust and reliable ML systems. Our study provides insights into how errors between different ML models are correlated. Through these insights, we aim to help organizations better assess the risks associated with deploying multiple models, particularly in critical or high-stakes environments. Furthermore, this work lays the groundwork for developing more effective risk management strategies and governance frameworks that can be employed to mitigate the cascading effects of correlated model failures. Ultimately, our paper contributes to a more comprehensive understanding of how ML models interact, while also highlighting the need for further research focused on enhancing the resilience and reliability of AI-driven applications. We now summarize the contributions of the current work.

\textbf{Current contributions:} We make the following contributions:
\begin{itemize}
    \item We  introduces a novel framework for systematically analyzing the correlations between errors in multiple ML models. The proposed framework defines the correlation mathematically by identifying the underlying random variables and sources of uncertainty that influence model performance. 
    \item We outline specific scenarios in which such correlations are likely to emerge, focusing on common practical situations where ML models are deployed in parallel or sequentially. 
    \item We empirically investigate how the correlations in model errors manifest and the implications they have for system-level performance. 
\end{itemize}

\section{Related Work}
\textbf{Ensemble Learning:} 
Ensemble learning intends to combine models to produce more accurate and reliable predictions. Similar to our work, the error correlation between these models is of interest. The greater the extent of error independence between the models, the better the performance of the ensemble \cite{Breiman1996BaggingP,Kuncheva2003MeasuresOD}. Several measures have been introduced in the literature to quantify this concept of diversity, including pairwise measures such as the Q-statistic, correlation coefficient, and disagreement measure \cite{Kuncheva2003MeasuresOD, disagreement}. Non-pairwise measures are also widely employed, often based on the average variation in models' losses \cite{Wood2023AUT}. In addition to improving performance, research shows that diverse ensembles are robust against adversarial attacks \cite{Biggio2011, Pang2019ImprovingAR} and can counteract covariate shift \cite{sinha2021dibs}. These benefits have also motivated the investigation into understanding and quantifying model diversity and correlations in this paper. 

\textbf{Error variance analysis:} Model selection in ML is often done by computing errors either on a hold-out set or via cross-validation. For reliable model selection, an accurate grasp on the variance of these errors and its estimation is paramount. Several previous works have developed theoretical framework to analyze this variance. Already in \cite{PatternRecog}, authors derived bounds for the variance of cross-validation for nearest neighbor type algorithms. The difficulty of accurately estimating this variance has been studied extensively in \cite{StatisticalTestsDietterich,InferenceGeneralizationError,NoUnbiasedYoshau}. Furthermore, the authors in \cite{InferenceGeneralizationError} investigated the theoretical and practical merits of different variance estimation techniques for cross-validation. We refer to \cite{BiasErrorEstimation,ResidualVariance,VarianceAnalysisCV,ProblemCV} for additional works on the same topic.

\textbf{Comparison to current work:} While the ensemble model literature primarily emphasizes leveraging diversity to enhance an ensemble model's predictive performance and robustness, this study focuses on identifying scenarios where models exhibit significant correlations. Recognizing such scenarios is crucial for developing strategies to mitigate risks, such as widespread failures caused by adversarial attacks and data drift. By studying these correlations, we aim to establish a robust framework for risk management, which would help in enhancing the reliability and resilience of machine learning models in real-world applications.

The aforementioned error variance analysis has focused primarily on errors resulting from one model. However, the current work is concerned with identifying scenarios where errors from \textit{different} models could be correlated, and numerically investigating these correlations for state-of-the-art models. To the best of our knowledge, ours is the only work that examines such correlations. We argue that our framework that considers multiple models deployed simultaneously is closer to real life scenarios where multiple models are deployed for similar applications, resulting in correlations. 

\section{Correlation Scenarios} \label{sec:scenario}
Below, we elaborate on the three relevant scenarios we identified where the errors between different models could be correlated. \autoref{table: scenario summary} summarizes these three scenarios.
\begin{enumerate}
    \item \textbf{Scenario 1: Different model architecture but same dataset} With the advent of public datasets, same problem might be solved by different technology providers using different model architectures but the same dataset. For instance, in computer vision, technology providers may train their own convolutional neural network model (ResNet, VGG, DenseNet, etc.) using the same public datasets like ImageNet \cite{imagenet} and Microsoft COCO datasets \cite{coco}. This similarity of the underlying dataset is likely to result in correlated errors.
    \item \textbf{Scenario 2: Overlapping features but same model architecture and same dataset} When models are trained on datasets with similar but not identical features—such as datasets with overlapping covariates—their predictions and associated errors can exhibit correlations. This scenario is particularly common in tabular datasets, where different organizations utilize same public datasets to build models containing a shared subset of features, with only minor differences in additional covariates---battery health prediction, for instance, often uses the same public dataset but different overlapping features \cite{BatteryHu,batteryDas}. If the overlapping features are highly predictive of the outcome, the differences between the models' predictions become minimal. Consequently, the models are likely to exhibit consistent behavior, leading to highly correlated error terms.
    \item \textbf{Scenario-3: Different fine-tuned models, different datasets but same foundation model} Different machine learning models can be built upon the same foundation model \cite{kamath2024large}. In such case, the weights of foundation models influence the fine-tuned models, potentially leading to correlations in predictions across different fine-tuning tasks. Further details can be found in our experiment section below.
\end{enumerate}

\begin{table}[htbp]
    \centering
    \begin{adjustbox}{width=\columnwidth}
    \begin{tabular}{c|c|c|c|c}
    \hline
    \textbf{Scenario} & \textbf{Overlapping features} & \textbf{Same Dataset} & \textbf{Same model architecture} & \textbf{Same foundation model}\\
    \hline
        \textbf{Scenario-1} & \textcolor{green}{\cmark} (all) & \textcolor{green}{\cmark} & \textcolor{red}{\xmark} & N/A\\
        \hline
        \textbf{Scenario-2} & \textcolor{green}{\cmark} (some) & \textcolor{green}{\cmark} & \textcolor{green}{\cmark} & N/A\\
        \hline
        \textbf{Scenario-3} & \textcolor{red}{\xmark} & \textcolor{red}{\xmark} & \textcolor{red}{\xmark} & \textcolor{green}{\cmark}\\
        \hline
    \end{tabular}%
    \end{adjustbox}
    \caption{Summary of Scenarios.}
    \label{table: scenario summary}
\end{table}

\subsection{Definition of error correlation}
Before defining the error correlations, we first assign notations to training dataset, errors and models. Let $\{(X_i, Y_i)\}_{i=1}^n$ be the training data, $\hat{f}(\cdot)$ be the model trained in the training data. Let $X_{n+1}$ be a future input to the trained ML model $\hat{f}(\cdot)$, and $Y_{n+1}$ be the ground-truth value of the outcome for $X_{n+1}$, the error term is defined as the difference between the ground-truth value and model predicted value, i.e.,
\begin{equation}
    \epsilon(Y_{n+1},\hat{f}(X_{n+1}))= Y_{n+1}-\hat{f}(X_{n+1}).\label{def: error_reg}
\end{equation}
Futhermore, for classification, we define the error as
\begin{equation}
    \epsilon(Y_{n+1},\hat{f}(X_{n+1}))= \mathds{1}_{Y_{n+1} \neq \hat{f}(X_{n+1})}.\label{def: error_cla}
\end{equation}
Notice that the error term is a function of the test data ($X_{n+1}$,$Y_{n+1}$) and the trained model $\hat{f}$. Therefore, similar models may result in similar error terms on the same input test data. This results in errors being correlated across different models. 

Using the above notation, the correlation (Corr) of error between two models is defined as below.
\begin{definition}[Correlation of error terms across models]\label{def:corr_error}
The correlation of error terms between two models \(\hat{f}_1\) and \(\hat{f}_2\) is given by    
\begin{equation}
    \rho_{\hat{f}_1, \hat{f}_2} = \text{Corr}\left(\epsilon(Y_{n+1}, \hat{f}_1(X_{n+1})), \epsilon(Y_{n+1}, \hat{f}_2(X_{n+1}))\right).\label{eq:rho_error}
\end{equation}
\end{definition}
Note that the randomness in above expression comes from the training and test datasets. 

The above definition suffices for Scenario-1 and Scenario-2. For Scenario-3, we would like to study how the aggregated performance of a fine-tuned model over a test-set correlates to the aggregate performance of another fine-tuned model. Any performance metric (like precision, recall, etc.) that aggregates the model's performance over a test set would suffice here. Let $E(\hat f,\mathcal{Z})$ represent such a performance metric that computes the performance of a model $\hat f$ over a testset $\mathcal{Z}$. Then, we define the correlation as below.
\begin{definition}[Correlations of performance between fine-tuned models]\label{def:corr_finetune}
    Let \(F\) be a pre-trained foundation model, and let \(\hat{f}_1(F)\) and \(\hat{f}_2(F)\) be two fine-tuned models based on \(F\). Let
    \(\mathcal{Z}_1\) and \(\mathcal{Z}_2\) be the test datasets for \(\hat{f}_1(F)\) and \(\hat{f}_2(F)\), respectively. The correlation of the performance is then given by:
    \begin{equation}
        \Pi_{\hat{f}_1(F), \hat{f}_2(F)} = \text{Corr}\left(E(\hat{f}_1(F), \mathcal{Z}_1), E(\hat{f}_2(F), \mathcal{Z}_2)\right).\label{eq:rho_finetune}
    \end{equation}
\end{definition}
In this work, for simplicity, we consider the average error, which reads
\begin{align}
    E_{\text{avg}}(\hat f, \mathcal{Z})=\frac{1}{n}\sum_{i=1}^n\epsilon(Y_i, \hat{f}(X_i)), \label{average error}
\end{align}
where $\hat f$ and $\mathcal{Z}$ represent the model and the testset, respectively.
\begin{remark}[Closed-form solution]
Since the distributions for test and training datasets are often intractable and may not have closed-form representations, obtaining an explicit analytical solution for correlation is difficult. Hence, we later undertake an empirical approach that involves using various example datasets to simulate the scenarios of interest, allowing us to estimate the correlations in practice.    
\end{remark}
\begin{remark}[Correlation coefficients]
To study the correlations, we use correlation coefficients. One of the coefficients we consider is the Pearson's correlation coefficient \cite{Benesty2009}. Since Pearson's coefficient is inadequate in capturing non-linear relationships among categorical variables, we also consider the $\phi_K$ measure \cite{baak2019corr_categorical}. $\phi_K$ is a versatile correlation measure that works across variable types (numerical, categorical, or mixed) and can capture both linear and non-linear dependencies. We use Pearson's correlation coefficient to compute the correlation between continuous variables and $\phi_K$ for categorical variables. Therefore, given the definition of error in \ref{def: error_reg} and \ref{def: error_cla}, we use Pearson's coefficient and $\phi_K$ for regression and classification, respectively.
\end{remark}

\section{Experiments}\label{sec:experiment}
We conduct experiments to quantify model correlations in the scenarios described in \autoref{sec:scenario}. The code is available online at \href{https://github.com/YuanyuanLi96/Corr_ML}{this repository}. 

\subsection{Datasets}
Throughout the experiments, we use 1 tabular dataset (California Housing \cite{KELLEYPACE1997291}), 4 image datasets (CIFAR-10 \cite{Krizhevsky2009LearningML}, EUROSAT \cite{helber2019eurosat}, MNIST \cite{lecun2010mnist}, Fashion-MNIST \cite{fashion}) and 4 text datasets (financial\_phrasebank \cite{Malo2014GoodDO}, twitter-financial-news-sentiment \cite{twiteer_financial}, emotion-balanced \cite{saravia-etal-2018-carer}, ag\_news \cite{Zhang2015CharacterlevelCN}) for the classification tasks. The tabular and images datasets are available in the Sklearn \cite{pedregosa2011scikit} and Tensorflow \cite{abadi2016tensorflow} Python packages, and the text datasets are downloaded from Hugging Face \cite{hf_data}. These datasets collectively offer comprehensive evaluation benchmarks, facilitating the training and assessment of machine learning models across a spectrum of tasks. We provide the descriptions of all the datasets in \autoref{tab:datasets_with_data_type}.

As we focus on quantifying the error correlations across models, the accuracy of the models are not optimized in our experiments to allow faster computations and more variations in the errors. Therefore we use random sampling to get the subsets of datasets for training and testing when the original datasets are large. For regression task, the data size for training and testing are 16512 and 4128 (i.e., 80-20 random split). For image classification, we sample $n_{\text{train}}=2000$ training and $n_{\text{test}}=500$ testing data from each datasets. For text classification models, we finetune the Large Language Models on the 4 text datasets, with $n_{\text{train}}=n_{\text{test}}=300 \times K$, where $K$ is the number of classes in the labels of each dataset.

\begin{table}[htbp]
    \centering
    \begin{adjustbox}{width=\columnwidth}
    \begin{tabular}{lll}
    \toprule
    \textbf{Dataset Name} & \textbf{Purpose} & \textbf{Outcomes}  \\
    \midrule
    California Housing \cite{KELLEYPACE1997291} & Housing Prices (Regression) & Median house value for districts \\
    CIFAR-10 \cite{Krizhevsky2009LearningML}& Common Objects (Classification) & 10 classes (e.g., airplane, cars) \\
    EUROSAT \cite{helber2019eurosat} & Satellite Images of Land (Classification) & 10 classes (e.g., river, forest)  \\
    MNIST \cite{lecun2010mnist} & Handwritten Digits (Classification) & 10 classes (digits 0-9) \\
    FASHION \cite{fashion} & Fashion Items (Classification) & 10 classes (e.g., T-shirt, trouser) \\
    financial\_phrasebank \cite{Malo2014GoodDO} & Financial News (Classification) & Positive, Negative, Neutral  \\
    twitter-financial-news-sentiment \cite{twiteer_financial} & Finance-related Tweets (Classification) & Bullish, Bearish, Neutral \\
    emotion-balanced \cite{saravia-etal-2018-carer} & English Twitter messages (Classification) & Joy, Sadness, Anger, Fear, Love, Surprise, Neutral \\
    ag\_news \cite{Zhang2015CharacterlevelCN} & News articles (Classification) & World, Sports, Business, Sci/Tech \\
    \bottomrule
    \end{tabular}%
    \end{adjustbox}
    \vskip 0.2cm
    \caption{Description of Datasets.}
    \label{tab:datasets_with_data_type}
\end{table}

\subsection{Models}

On the tabular dataset, we perform regression using various types of models: linear regression, random forest, XGBoost, generalized additive model and neural network models. For image classification tasks, we also create models using different algorithms including logistic regression, random forest, XGBoost, neural network model, and convolutional neural network models. In the experiments that involve foundation models, we download pre-trained foundation models from Tensorflow library \cite{abadi2016tensorflow} or Hugging Face \cite{hf_models}. Fine-tuning of foundational image classification models is done by adding one trainable dense layer with 64 nodes on top of the foundation models. For large language models, we apply Parameter-Efficient Fine-Tuning (PEFT) \cite{peft}, including techniques such as Low-Rank Adaptation on all linear layers. The descriptions of all fully trained models and the pre-trained foundation models are summarized in \autoref{tab:combined_models}. All implementation is done in Python.

\begin{table}[htbp]
    \centering
    \begin{adjustbox}{width=\columnwidth}
    \begin{tabular}{ll}
    \toprule
    \textbf{Model} & \textbf{Number of parameters} \\
    \midrule
    linear_regression \cite{Free:2005}& Number of input features $p$\\
    logistic_regression \cite{cox1958regression}& $p$ \\
    random_forest \cite{Breiman2001RandomF}& depends on number of trees and nodes\\
    xgboost \cite{chen2016xgboost}& depends on number of trees and nodes \\
    gam \cite{gam}& $p$ $\times$ number of parameters in smooth functions \\
    NN1 \cite{Rumelhart1986LearningRB}& 1 hidden layer with 64 neurons   \\
    NN2 \cite{Rumelhart1986LearningRB}& 2 hidden layers with 64 and 32 neurons\\
    CNN1 \cite{deeplearning}& 1 hidden convolutional layer with 64 neurons   \\
    CNN2 \cite{deeplearning}& 2 hidden layers with 64 and 32 neurons \\
    ResNet50 \cite{he2016deep}& 25.6M\\
    ResNet101 \cite{he2016deep}& 44.5M \\
    VGG16 \cite{simonyan2014very}& 138M \\
    VGG19 \cite{simonyan2014very}& 144M \\
    DenseNet121 \cite{huang2017densely}& 7.9M\\
    DenseNet169 \cite{huang2017densely}& 14M\\
    MobileNet \cite{howard2017mobilenets}& 4.2M \\
    MobileNetV2 \cite{sandler2018mobilenetv2}& 3.4M \\
    Mistral-7B-v0.3 \cite{mistral7b}& 7B \\
    Meta-LLaMA-3-8B \cite{llama3}& 8B \\
    Qwen2-7B \cite{yang2024qwen2}& 7B\\
    Llama-2-7b-hf \cite{llama2}& 7B  \\
    aya-23-8B \cite{aryabumi2024aya}& 8B  \\
    falcon-7b \cite{refinedweb}& 7B \\
    bloom-7b1 \cite{bloom7b1}& 7.1B \\
    phi-2 \cite{javaheripi2023phi}&2B\\
    \bottomrule
    \end{tabular}%
    \end{adjustbox}
    \vskip 0.2cm
    \caption{Description of Models.}
    \label{tab:combined_models}
\end{table}

\subsection{Results}
\subsubsection{(Scenario-1) Similar algorithms lead to similar errors}
\begin{figure}[htbp]
  \centering
  \includegraphics[width=0.4\textwidth]{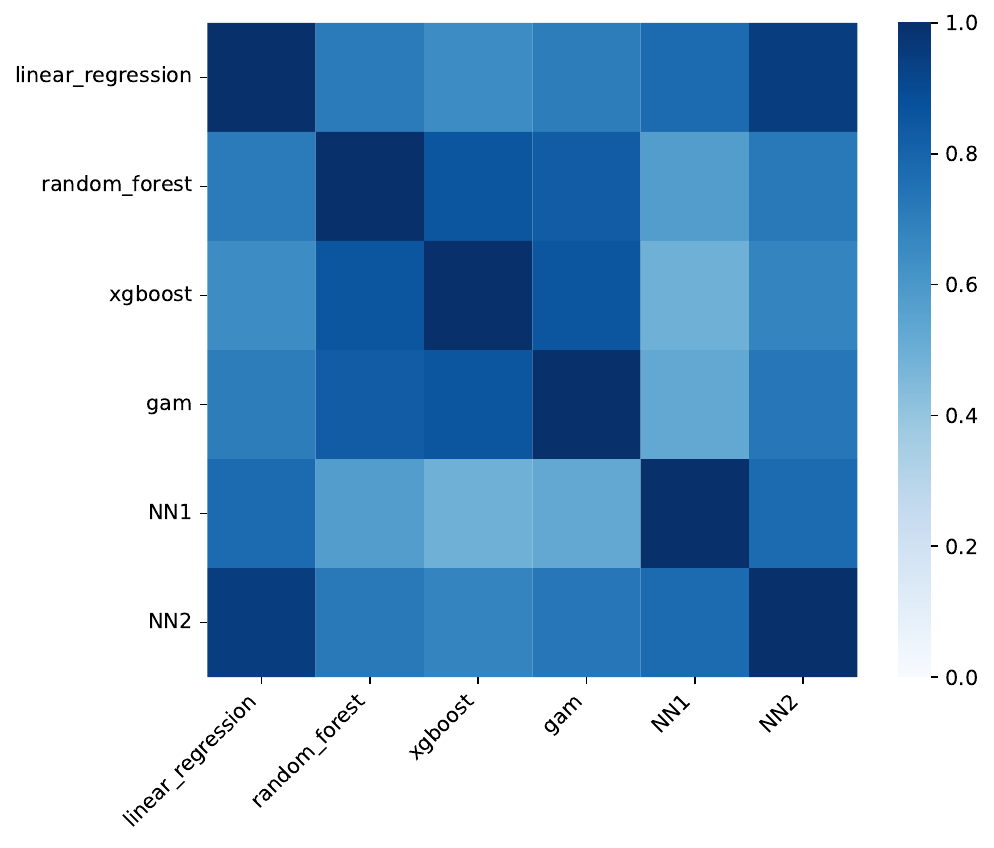}
  \caption{Pearson correlations between errors of different models on the same data (Scenario 1; tabular data). The errors are computed on California Housing dataset. Darker colors represent higher correlations.}
  \label{fig:fully_trained_tab}
\end{figure}

For the dataset presented in \autoref{tab:datasets_with_data_type}, we train and evaluate the various models listed in \autoref{tab:combined_models}. Subsequently, we calculate the testing errors of each model and compute their pairwise correlation coefficients. Specifically, for the California Housing dataset, we present the Pearson's correlation coefficient matrix of model errors in \autoref{fig:fully_trained_tab}, where the color scheme is designed to reflect the magnitude of the correlation—darker shades correspond to higher values. Our analysis reveals that all correlation coefficients are positive, indicating a positive correlation between the models' error terms. A correlation coefficient greater than 0.7 is generally considered indicative of a strong correlation \cite{strong_correlation}. From the figure, we observe that Random Forest, XGBoost, and the Generalized Additive Model (GAM) exhibit a strong correlation. Additionally, the two neural network models, NN1 and NN2, show a significant correlation with each other, as well as with the linear regression model. This suggests that similar model architectures tend to produce similar model predictions and error patterns.

\begin{figure}[htbp]
  \centering
  \includegraphics[width=0.4\textwidth]{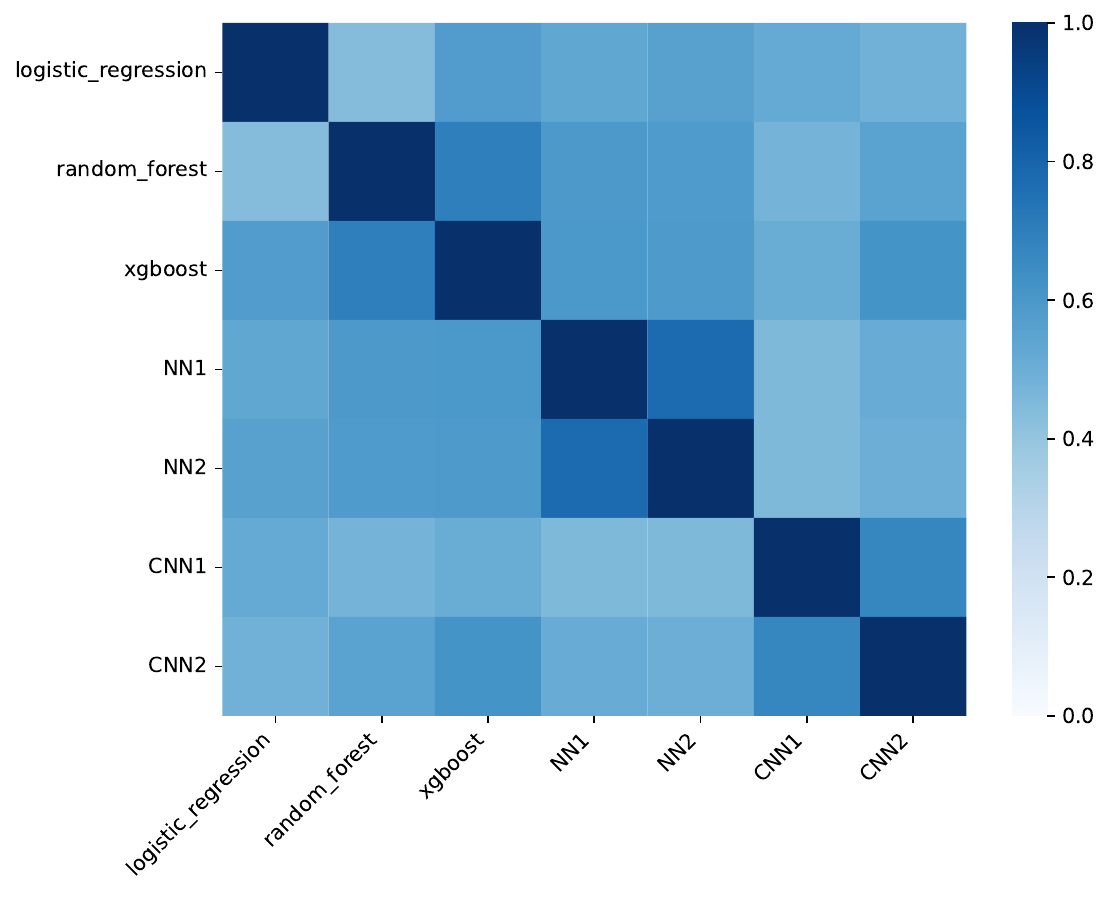}
  \caption{$\phi_K$ correlations between errors of different models on the same data (Scenario 1; image data). The errors are computed on CIFAR-10 dataset.}
  \label{fig:fully_trained_image}
\end{figure}

We present the correlations of image classification model errors on the CIFAR-10 dataset in \autoref{fig:fully_trained_image}. The correlation metric used is $\phi_K$ \cite{baak2019corr_categorical}, which is adapted to handle binary classification errors. As observed with regression models, the errors of the random forest and XGBoost models exhibit a high degree of correlation. In contrast, logistic regression shows weak to moderate correlations with all other models. The two neural network models, NN1 and NN2, are strongly correlated with each other. Similarly, CNN1 and CNN2 display a high degree of correlation but show only weak correlations with the other models.

\begin{figure}[htbp]
  \centering
  \includegraphics[width=0.4\textwidth]{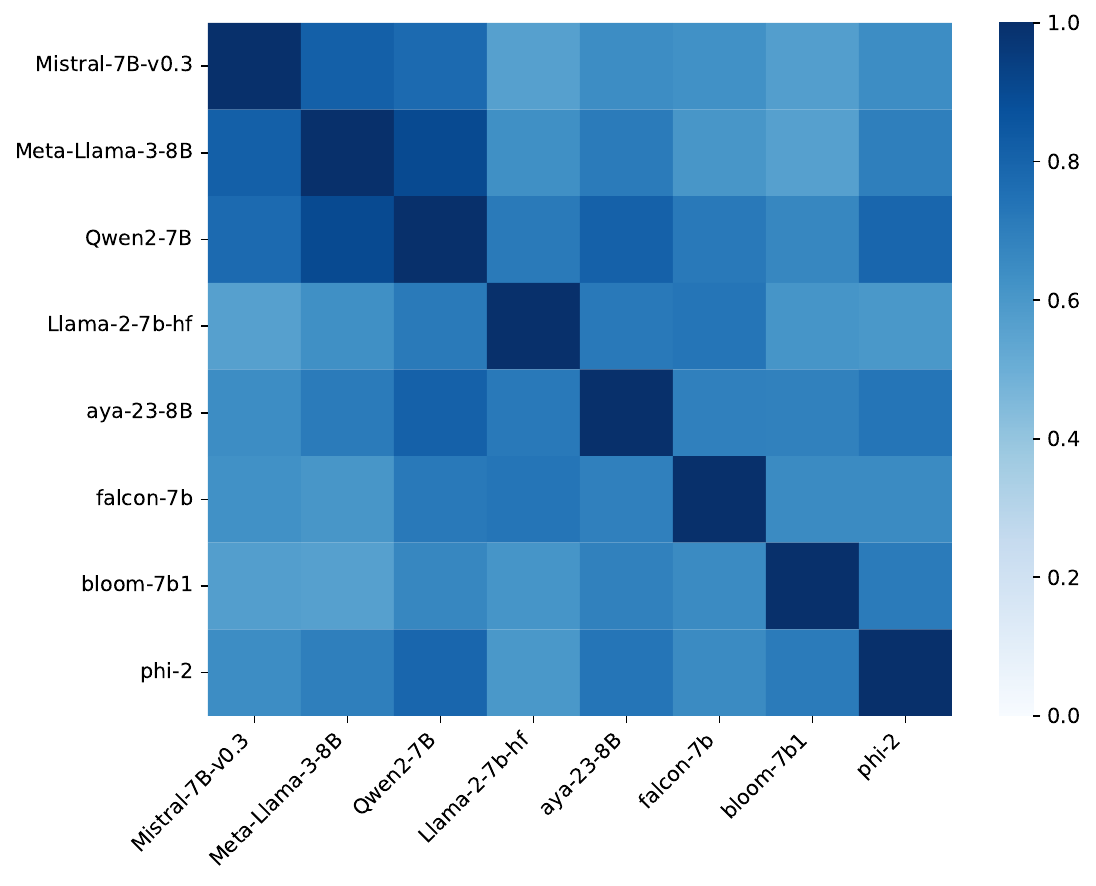}
  \caption{$\phi_K$ correlations between errors of different models on the same data (Scenario 1; text data). The errors are computed on financial\_phrasebank dataset.}
  \label{fig:cor_llm}
\end{figure}

For the text classification task, we fine-tune the large language models (LLMs) listed in \autoref{tab:combined_models} on the training split of the financial\_phrasebank dataset described in \autoref{tab:datasets_with_data_type}. Consistent with the image classification results, we use the $\phi_K$ correlation metric to measure the correlations between the errors of different text classification models. The corresponding correlation plot is shown in \autoref{fig:cor_llm}. Interestingly, we observe more widespread correlations across the LLMs compared to those of the image classification models in \autoref{fig:fully_trained_image}. This finding aligns with our expectations, as all LLMs are based on foundational transformer architectures, which share greater similarity in their model structures and therefore exhibit more consistent error patterns.

\subsubsection{(Scenario-2) Similar features lead to similar errors}
\begin{figure}[htbp]
  \centering
  \includegraphics[width=0.4\textwidth]{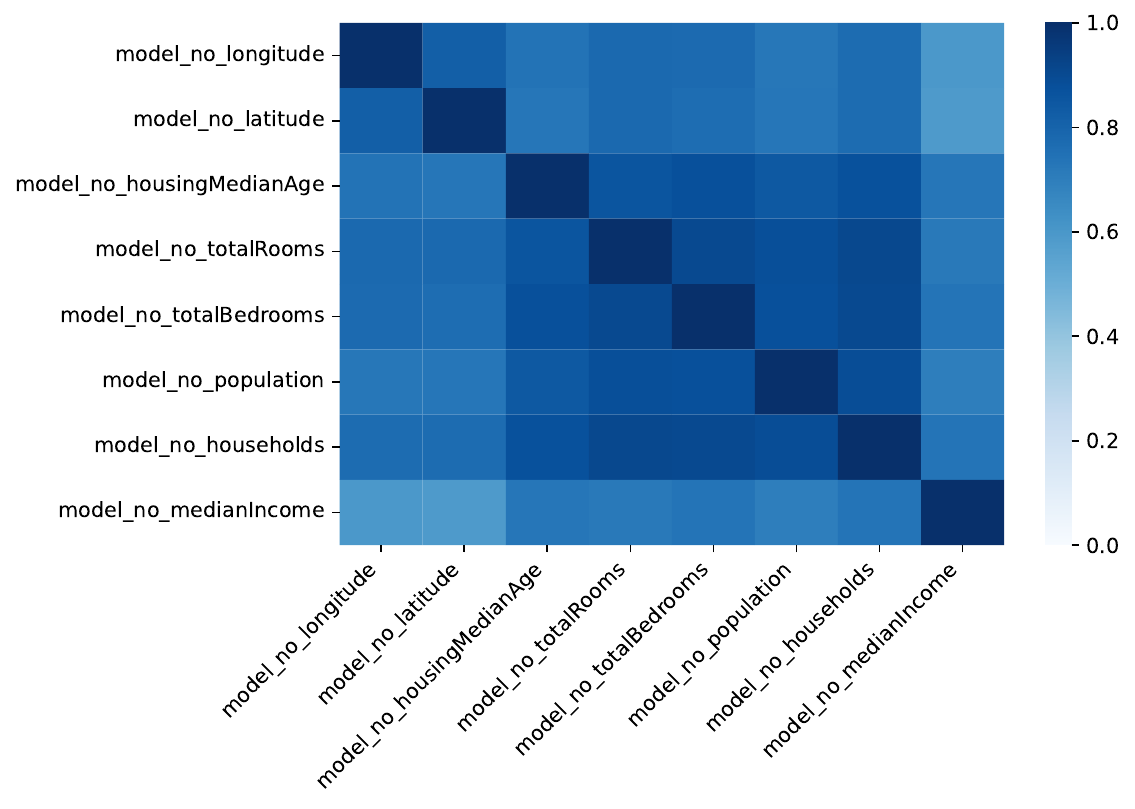}
  \caption{Pearson correlations between errors of different models with overlapping features (Scenario 2; tabular data) on California Housing dataset. }
  \label{fig:cor_overlap}
\end{figure}

We use the California Housing dataset and create 8 XGBoost models, each removing a different input variable from the original dataset. We then calculate the Pearson's correlation coefficients of the regression model errors. The results, presented in \autoref{fig:cor_overlap}, show that all models exhibit moderate to strong correlations with each other. A notable group of models, $\mathcal{M}_f$= \{model\_no\_housingMedianAge, 
model\_no\_totalRooms, \\
model\_no\_population, 
model\_no\_households\}, demonstrates strong correlations. This group corresponds to those models that remove features from the set \{housingMedianAge, totalRooms, population, households\}.

\begin{figure}[htbp]
  \centering
  \includegraphics[width=0.4\textwidth]{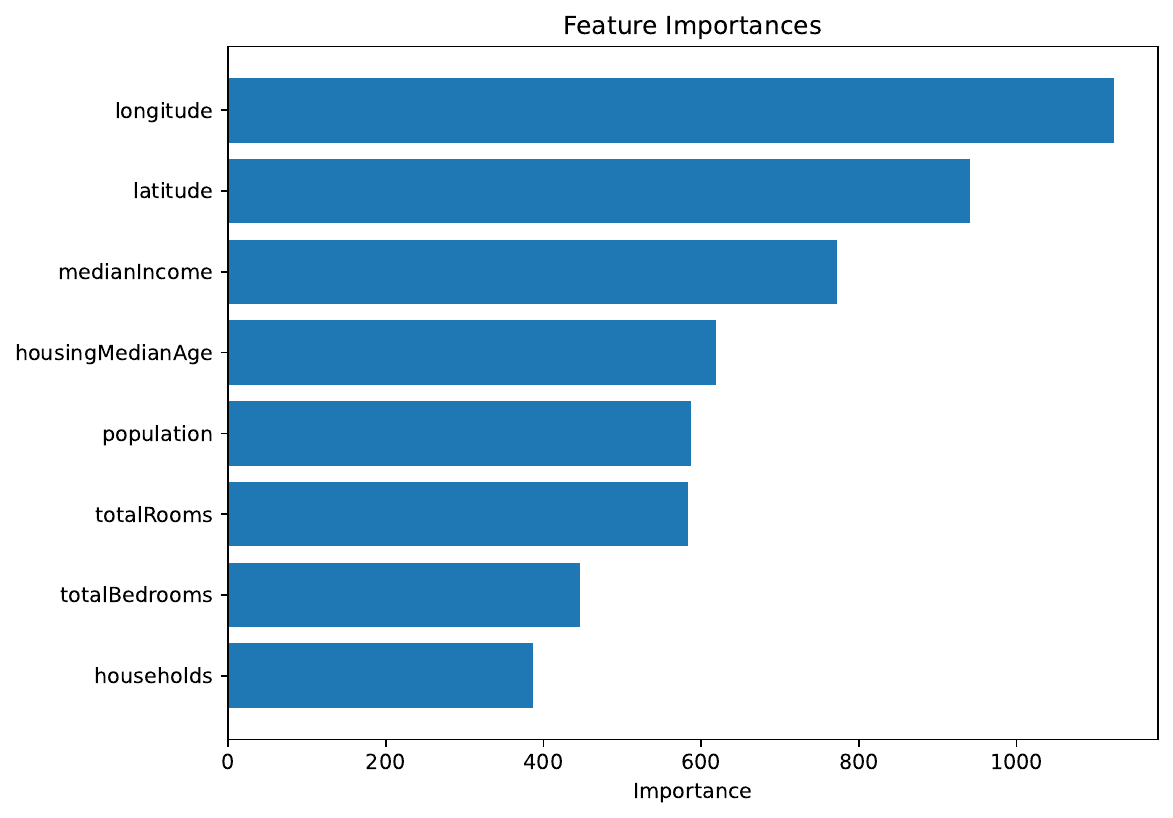}
  \caption{Feature importance of XGBoost model on California Housing dataset. }
  \label{fig:feature_importance}
\end{figure}

As predicted in \autoref{sec:scenario}, when the overlapping features among models are highly predictive of the outcome, the differences between the models become negligible, leading to similar error patterns. To validate this projection, we calculate the predictive power of all features in the XGBoost model using feature importance, as shown in \autoref{fig:feature_importance}. The analysis reveals that the most predictive features—${\text{longitude}, \text{latitude}, \text{medianIncome}}$—are also the overlapping features of the models exhibiting strong correlations, $\mathcal{M_s}$. This finding aligns with and confirms our projection.

\subsubsection{(Scenario-3) Errors in foundation models propagate downstream through fintuning}
\begin{figure}[htbp]
    \centering
    \begin{subfigure}[b]{0.4\textwidth}
        \centering
        \includegraphics[width=\textwidth]{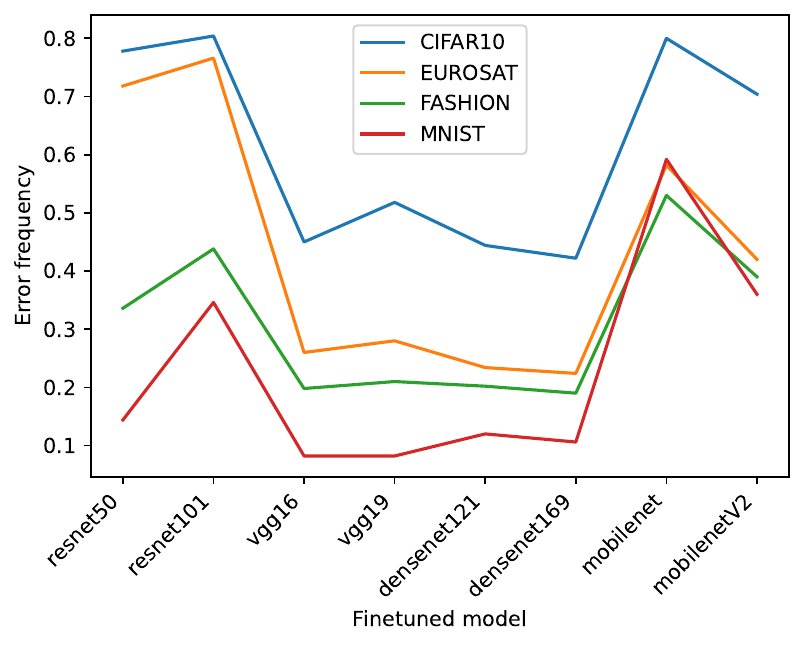}
        \caption{Error frequencies of fine-tuned foundation models on different datasets.}
        \label{fig:correlation_finetuneA}
    \end{subfigure}
    \hfill
    \begin{subfigure}[b]{0.3\textwidth}
        \centering
        \includegraphics[width=\textwidth]{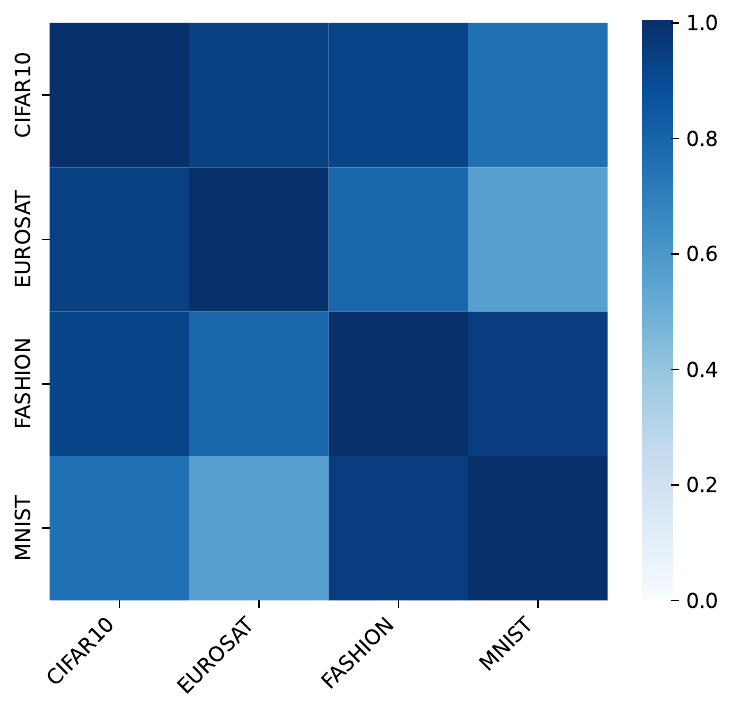}
        \caption{Pearson correlation between error frequencies of fine-tuned models on different datasets.}
        \label{fig:correlation_finetuneB}
    \end{subfigure}
    \caption{Model performance and their correlations on image finetuning datasets (Scenario 3; image data).}
    \label{fig: sce3_image}
\end{figure}

\begin{figure}[htbp]
    \centering
    \begin{subfigure}[b]{0.4\textwidth}
        \centering
        \includegraphics[width=\textwidth]{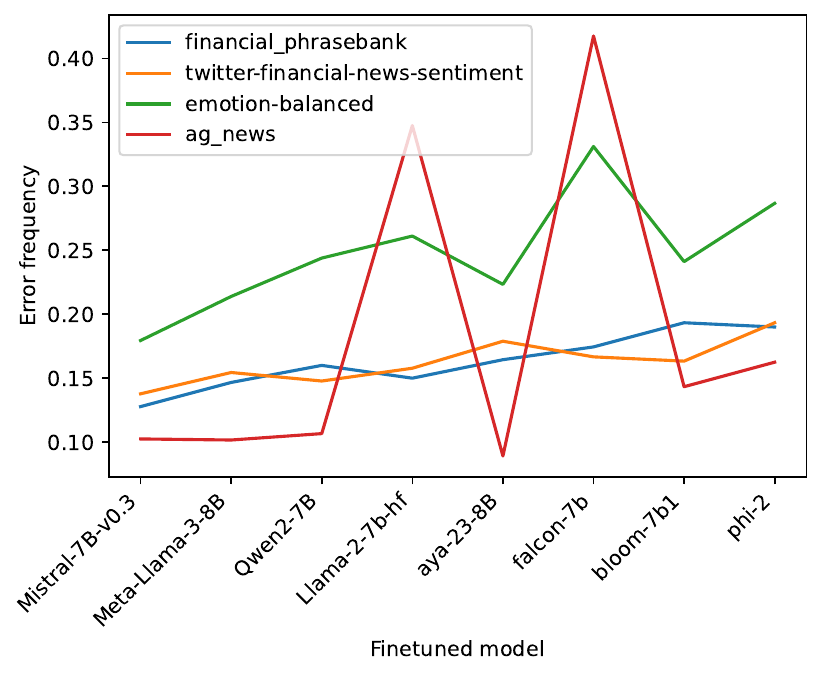}
        \caption{Error frequencies of fine-tuned foundation models on different datasets.}
        \label{fig:err_finetune}
    \end{subfigure}
    \hfill
    \begin{subfigure}[b]{0.4\textwidth}
        \centering
        \includegraphics[width=\textwidth]{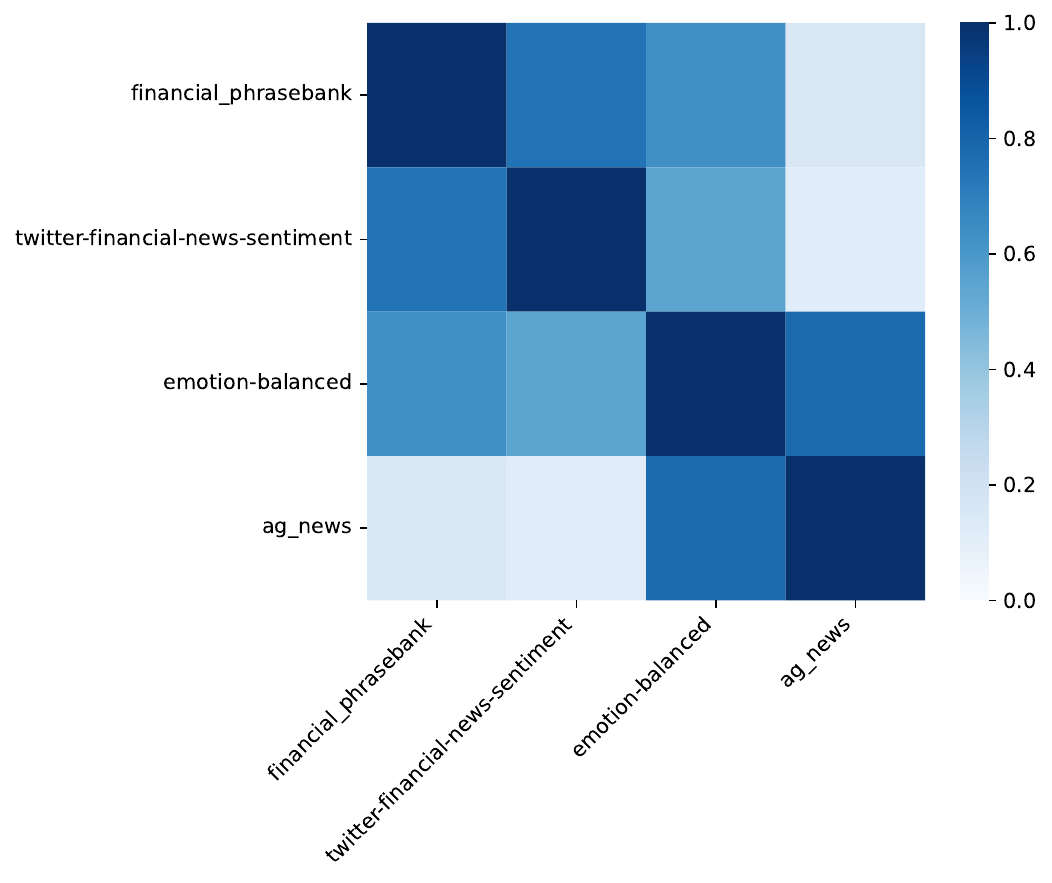}
        \caption{Pearson correlation between error frequencies of fine-tuned models on different datasets.}
        \label{fig:corr_finetune}
    \end{subfigure}
    \caption{Model performance and their correlations on text finetuning datasets (Scenario 3; text data).}
    \label{fig: sce3_llm}
\end{figure}

To evaluate the impact of foundational models on model correlations, we fine-tune the foundation models listed in \autoref{tab:combined_models} on the image and text datasets described in \autoref{tab:datasets_with_data_type}. We analyze the dependencies of model performance based on average error frequencies (see \autoref{average error}) and present the results in \autoref{fig: sce3_image} and \autoref{fig: sce3_llm}. In \autoref{fig:correlation_finetuneA}, the error frequencies across different datasets (represented by distinct colored lines) exhibit similar trends as the foundational models vary. For instance, error frequencies are consistently lower when using VGG16 or VGG19 compared to ResNet50 or ResNet101, regardless of the dataset. This observation suggests a high degree of correlation among the performance of fine-tuned models across datasets. We further compute the Pearson correlation coefficients, and the resulting correlation matrix, presented in \autoref{fig:correlation_finetuneB}, corroborates these findings.

For fine-tuned text classification models, performance on datasets such as financial\_phrasebank, financial\_sentiment, and emotion-balanced data varies consistently across different foundational models. In contrast, performance on the ag\_news dataset demonstrates greater variability when switching foundational models. This aligns with our understanding that news categorization represents a distinct use case compared to emotion detection, leading to lower correlations among fine-tuned models derived from the same foundational model. The Pearson correlation values across fine-tuned models, shown in \autoref{fig:corr_finetune}, reflect the expected magnitudes of these correlations.

\section{Conclusions}\label{sec:conclusions}
We propose a framework to analyze the correlations among machine learning models under three specific scenarios, supported by empirical studies. Our findings reveal that similar machine learning algorithms trained on the same dataset tend to produce highly correlated error terms, indicating potential risks of simultaneous failures due to algorithmic similarity. Models built using highly predictive overlapping features also exhibit similar error patterns, as the dominance of these features minimizes differences between models. Furthermore, models based on the same foundational architecture but fine-tuned on different datasets demonstrate correlated performance metrics, particularly when the datasets are similar, suggesting the potential for cascading failures across various use cases of foundational models. By examining these correlations, this paper aims to inform researchers and practitioners in designing robust AI systems and implementing effective risk management strategies to mitigate the risks associated with correlated failures in machine learning models.

\section{Discussions and Drawbacks}
This paper is not intended to provide a comprehensive assessment of all possible model correlation scenarios encountered in real-life applications. Instead, it aims to serve as a starting point to encourage further research on quantifying model correlations and systematically evaluating the potential consequences of model failures. By highlighting key scenarios and their implications, we hope to inspire more in-depth investigations into the risks and challenges posed by correlated machine learning model failures in diverse contexts. 

Being empirical in nature, the conclusions we have drawn are limited to the datasets and the models we have considered. It would be desirable to derive bounds on the error correlations at least for some simple models and under distributional assumptions. Such bounds could further corroborate the results we have presented here. We intend to present such theoretical results in the future.  

\medskip
\small
\bibliography{bib_accumulation}
\bibliographystyle{unsrt}

\end{document}